\documentclass{article}
\usepackage[main, final]{neurips_2025}
\usepackage[utf8]{inputenc} 
\usepackage[T1]{fontenc}    
\usepackage{hyperref}       
\usepackage{url}            
\usepackage{booktabs}       
\usepackage{amsfonts}       
\usepackage{nicefrac}       
\usepackage{float}

\usepackage[table, dvipsnames]{xcolor}
\definecolor{lightgreen}{RGB}{220, 245, 220}

\usepackage{times}
\usepackage{latexsym}
\usepackage{array}
\usepackage{multicol}
\usepackage{tablefootnote}
\usepackage{threeparttable}
\usepackage{pifont}
\usepackage{lipsum}
\setlipsum{%
  par-before = \begingroup\color{gray},
  par-after = \endgroup
}

\usepackage{graphicx}
\usepackage{amsmath}
\usepackage{mdframed}
\usepackage{subcaption}

\usepackage{xspace}
\usepackage{kotex} 

\usepackage{listings}  

\definecolor{lightgreen}{RGB}{220, 245, 220}

\usepackage{adjustbox}
\usepackage[breakable]{tcolorbox}

\usepackage{multirow}
\usepackage{svg}
\usepackage{titlesec}

\usepackage{bbm}
\usepackage{bm}

\usepackage{todonotes}     

\usepackage{natbib}

\usepackage{tabularx}
\usepackage{makecell}
\usepackage{colortbl}

\definecolor{customblue}{HTML}{6BE2F6}

\newcommand{\hc}{HyperCLOVA\xspace}
\newcommand{\hcx}{\hc X\xspace}
\newcommand{\hcxtfull}{\hcx 32B Think\xspace}
\newcommand{\hcxtshort}{THINK\xspace}
\newcommand{\hcxofull}{\hcx 8B Omni\xspace}
\newcommand{\hcxoshort}{OMNI\xspace}

\title{\hcxofull}
\def\huggingface{\raisebox{-1.5pt}{\includegraphics[height=1.05em]{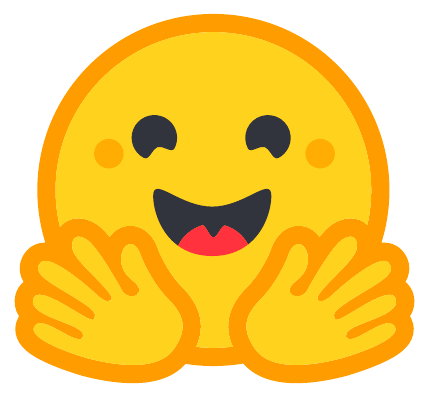}}}
\author{
NAVER Cloud\\
HyperCLOVA X Team\thanks{The complete list of contributors appears in the Contributions and Acknowledgments section. For correspondence, please contact \texttt{dl\_hcx\_technical\_report@navercorp.com}. \textcopyright\ 2025 NAVER Cloud HyperCLOVA X Team. All rights reserved.}\\
\small \href{https://huggingface.co/naver-hyperclovax/HyperCLOVAX-SEED-Omni-8B/tree/TAG-2025-12-31}{\huggingface\ \texttt{Huggingface Model Card}}
}

\begin{document}
\maketitle
\begin{abstract}
In this report, we present \hcxofull, the first any-to-any omnimodal model in the \hcx family that supports text, audio, and vision as both inputs and outputs. By consolidating multimodal understanding and generation into a single model rather than separate modality-specific pipelines, \hcxofull serves as an 8B-scale \textit{omni-pathfinding} point toward practical any-to-any omni assistants. At a high level, the model unifies modalities through a shared next-token prediction interface over an interleaved multimodal sequence, while vision and audio encoders inject continuous embeddings for fine-grained understanding and grounding. Empirical evaluations demonstrate competitive performance against comparably sized models across diverse input--output combinations spanning text, audio, and vision, in both Korean and English. We anticipate that the open-weight release of \hcxofull will support a wide range of research and deployment scenarios.
\end{abstract}
\section{Introduction} 
\begin{figure}[t]
\vskip -0.1in
\centering
\includegraphics[width=1.0\textwidth]{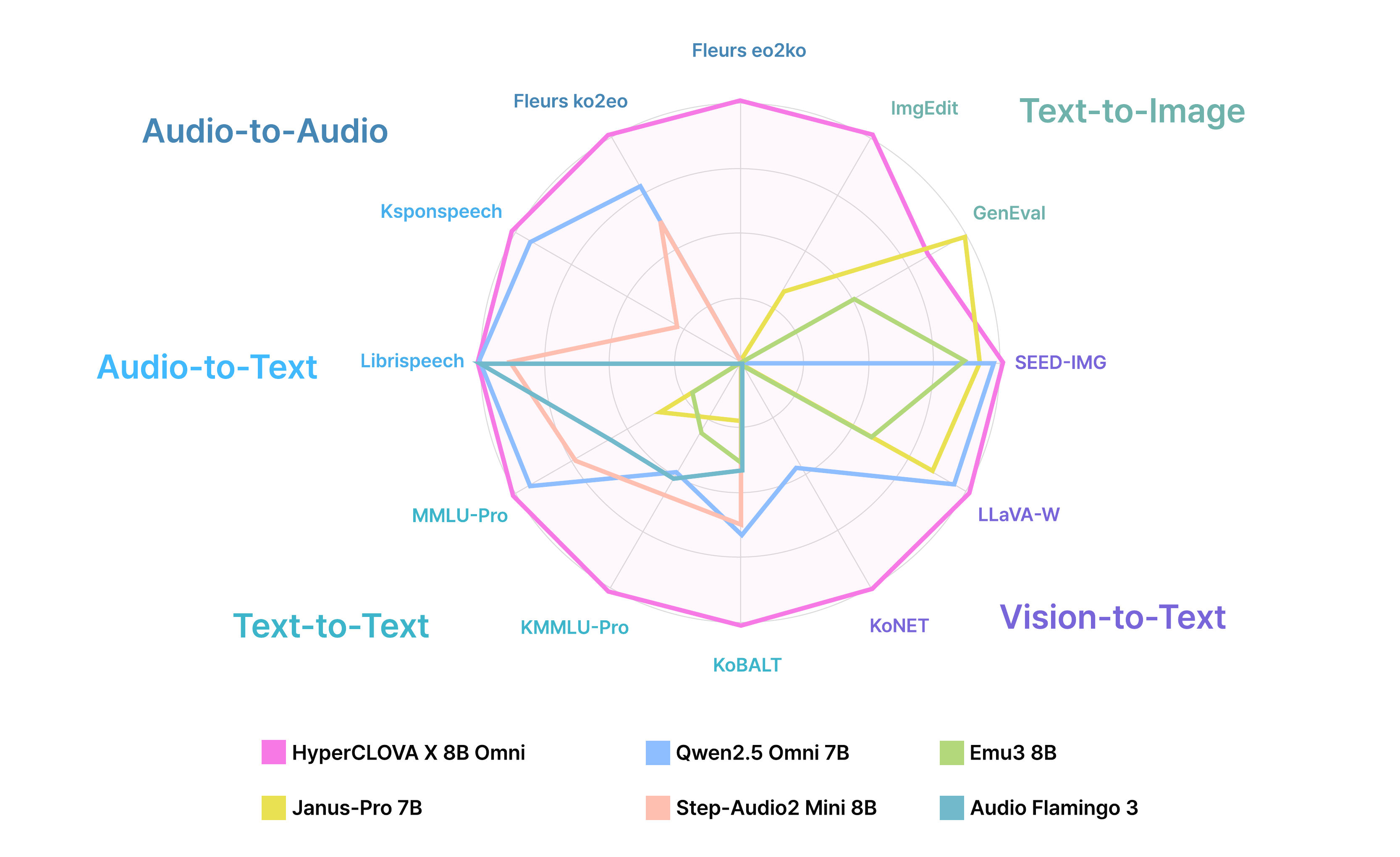}
\caption{Comparison of multimodal capabilities across text, vision, and audio for both generation and understanding tasks. The results highlight the unified end-to-end design of \hcxofull, which supports any-to-any multimodal understanding and generation within a single model.}
\label{fig:main_performa}
\vskip -0.07in
\end{figure}

The tight integration of AI systems into real-world contexts necessitates their ability to understand and generate across multiple modalities, such as text, audio, and vision. This requirement arises in part because specific applications inherently involve multimodal inputs and outputs. 
Moreover, human-generated text is projected to accumulate at a rate that cannot keep up with the rapid scaling of large language models~(LLMs;\citealt{10.5555/3692070.3694094}).
Even if it were the case, text alone cannot capture the full spectrum of multimodal dimensions of reality~\citep{DBLP:conf/icml/HuhC0I24,chen2025janus}.

One strategy for developing multimodal models extends existing LLMs by sequentially incorporating encoders and decoders for various modalities. While such modality extension enables a cost- and time-efficient transformation of a text-based model into a multimodal one, multimodal training often incurs catastrophic forgetting of knowledge within the LLM backbone~\citep{zhai2023investigating,palme,lee2025how,llavac}. This challenge calls for a joint training across multiple modalities in a unified framework.

In response, we introduce \hcxofull (\hcxoshort), an omnimodal model that supports text, audio, and vision modalities as both inputs and outputs, as shown in Figure~\ref{fig:main_performa}. \hcxoshort is a decoder-only Transformer jointly modeling an interleaved multimodal sequence of tokens and embeddings. Modality-specific tokens and embeddings share a common next-token prediction interface, thereby facilitating semantic composition across modalities. 

We compare the performance of \hcxoshort against that of comparably sized models on benchmarks spanning diverse combinations of input and output modalities, including text-to-text, vision-to-text, text-to-vision, speech-to-text, audio-to-text, and speech-to-speech. In addition, we present a human preference study on text-to-speech conversion. For most modality combinations, evaluations are carried out in both Korean and English to assess bilingual ability. The results demonstrate the competitive performance of \hcxoshort across the board, despite it being the only model capable of handling all combinations of input and output modalities. 

\hcxoshort is released as an open-weight model under a custom license that permits commercial use subject to specified conditions. Given its compact size and competitive performance across diverse input and output modalities, we present \hcxoshort as a valuable resource for academic and industry partners in both the Korean and global research community. 

\section{\hcxofull}

\begin{figure}[t]
\vskip -0.1in
\centering
\small
\includegraphics[width=1.0\textwidth]{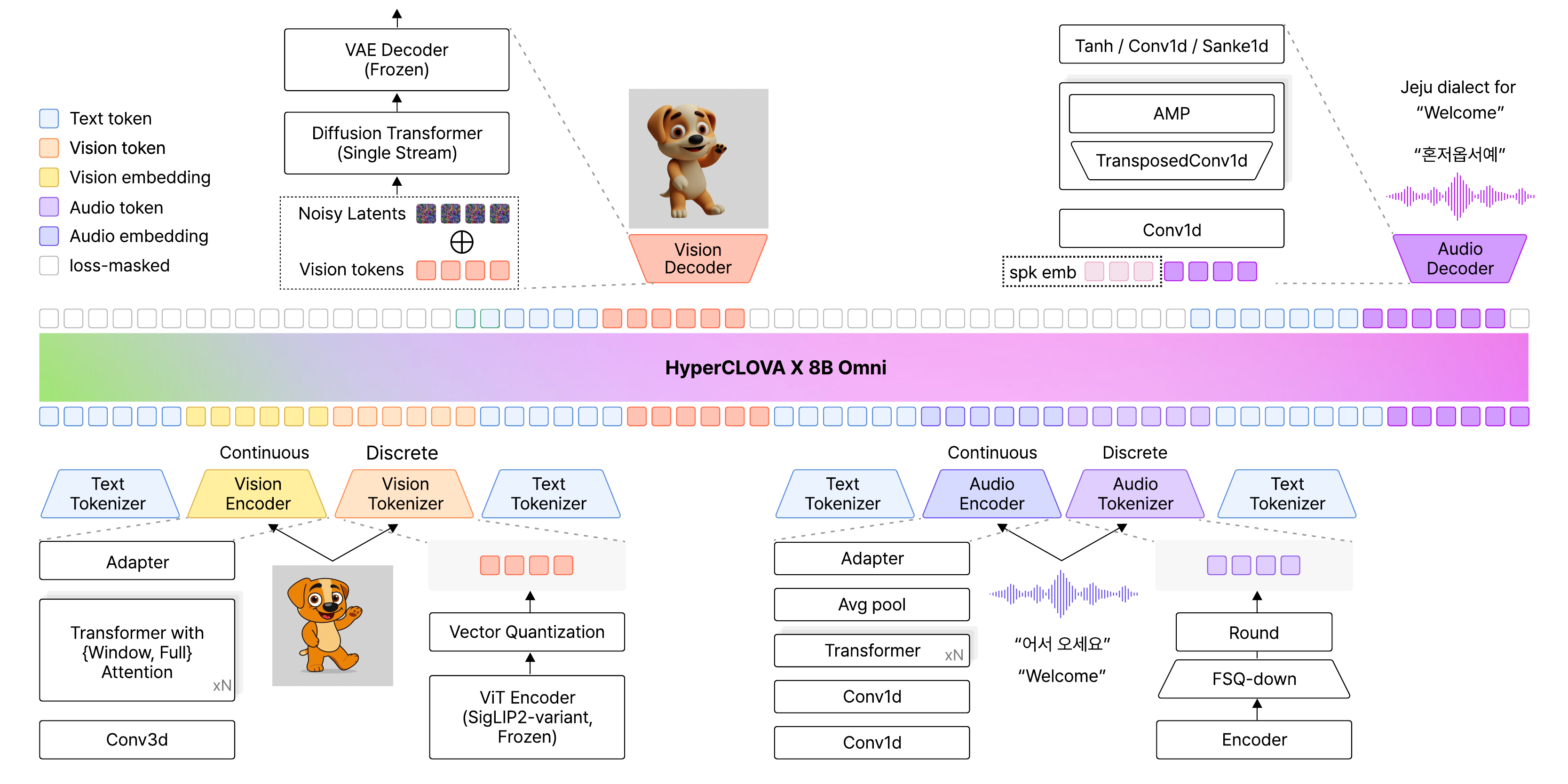}
\caption{Overall architecture of \hcxofull. Text, vision, and audio inputs are encoded into continuous embeddings and discrete tokens via modality-specific encoders and tokenizers, which are interleaved and jointly processed by a single decoder-only Transformer backbone. Modality-specific decoders reconstruct visual and auditory outputs from the shared sequence representations, enabling end-to-end any-to-any multimodal generation.
}
\label{fig:main_overview}
\vskip -0.05in
\end{figure}

\subsection{Design Motivation and Pathfinding}
Recent multimodal systems span a wide design space, ranging from late-fusion integration to modality-specific generation pipelines~\citep{team2025gemma, januspro, ai2025ming, Qwen2.5-Omni, Qwen2-Audio}. In our approach, the guiding hypothesis is that multimodal capabilities can be effectively realized when modality-specific tokens and embeddings share a common next-token prediction interface, enabling semantic composition across modalities. As illustrated in Figure~\ref{fig:main_overview}, text is represented as discrete tokens, while vision and audio are represented with both discrete tokens and continuous embeddings; these representations are interleaved and jointly processed by a single decoder-only Transformer backbone. 

We instantiate the backbone as a 36-layer auto-regressive Transformer with a hidden size of 4,096, closely following the architectural and implementation choices of \hcxtfull (\hcxtshort,~\citet{navercloudhyperclovaxteam2025hyperclovaxthinktechnical}). Following \hcxtshort, the text tokenization pipeline combines a morphology-preserving pretokenizer and a subword tokenizer and applies low-probability StoChasTok to mitigate token-boundary bias while preserving token efficiency. For subword tokenization, we adapt an English-centric tokenizer via a three-stage vocabulary modification, which significantly improves Korean token efficiency without degrading performance on English, code, or math tasks. 

Operationally, we unify multimodal generation by treating each modality tokenizer's discrete codebook entries as additional vocabulary items of the language model, thereby extending next-token prediction from text to a shared multimodal token space. For understanding and fine-grained grounding, we additionally attach modality encoders that produce continuous embeddings projected into the backbone embedding space. Modality decoders subsequently convert predicted non-text tokens into their native signal domains (pixels and waveforms).

The following sections provide detailed specifications of the vision and audio tokenizers/encoders and the associated decoders.

\subsection{Vision Modality}
\hcxoshort processes visual information through a synergistic integration of three components: a continuous vision encoder for perceptual understanding, a discrete semantic tokenizer for generative representation, and a diffusion-based decoder for pixel synthesis. This tripartite architecture is designed to natively handle interleaved multimodal sequences within a unified framework, in which each component plays a functional role.

First, a continuous vision encoder extracts dense features that are directly aligned with the LLM backbone to support overall vision understanding. Second, to support vision generation, \hcxoshort incorporates a vision tokenizer that quantizes visual features into discrete semantic tokens. This choice is closely tied to the auto-regressive (AR) nature of our Transformer backbone, which is inherently well-suited to modeling discrete tokens \citep{li_autoregressive_2024,deng_autoregressive_2024}. Unlike models such as Janus-Pro \citep{januspro} or Emu 3 \citep{emu3} that rely on low-level VAE-style tokenizers, our tokenizer operates at the semantic level to maximize cross-modal synergy with text embeddings \citep{zheng2025diffusion-RAE}—a critical advantage for our compact 8B backbone, where efficient and semantically aligned features are essential.

Finally, vision generation proceeds by decoding these discrete tokens into pixels using a diffusion-based vision decoder. Because semantic tokenization introduces unavoidable information loss by discarding fine-grained visual details, the diffusion model acts as a complementary component that stochastically recovers missing details. It synthesizes high-frequency textures and fine structures through a channel-concatenation-based architecture, which enables significantly faster convergence and supports near-native aspect ratios.

\paragraph{Encoder.}
Architecturally, the vision-understanding component of \hcxoshort follows \hcxtshort, adopting the Vision Transformer (ViT) architecture from Qwen2.5-VL~\citep{bai2025qwen2} for unified image and video modeling. For architectural stability, we utilize a streamlined linear adapter to align visual features with the LLM backbone~\citep{liu2023visual}. A primary design objective is computational efficiency; by optimizing the visual token allocation, we reduce training costs by approximately 53\% in GPU-hours compared to standard settings. Static images and 120-frame videos are compressed into efficient budgets of 3K and 11K tokens, respectively. Notably, the encoder remains unfrozen throughout training to establish Korean-centric multimodal capabilities, essential for internalizing Korean-specific visual contexts, cultural landmarks, and high-density OCR.

\begin{figure}
\vskip -0.1in
  \centering
  \includegraphics[width=0.9\linewidth]{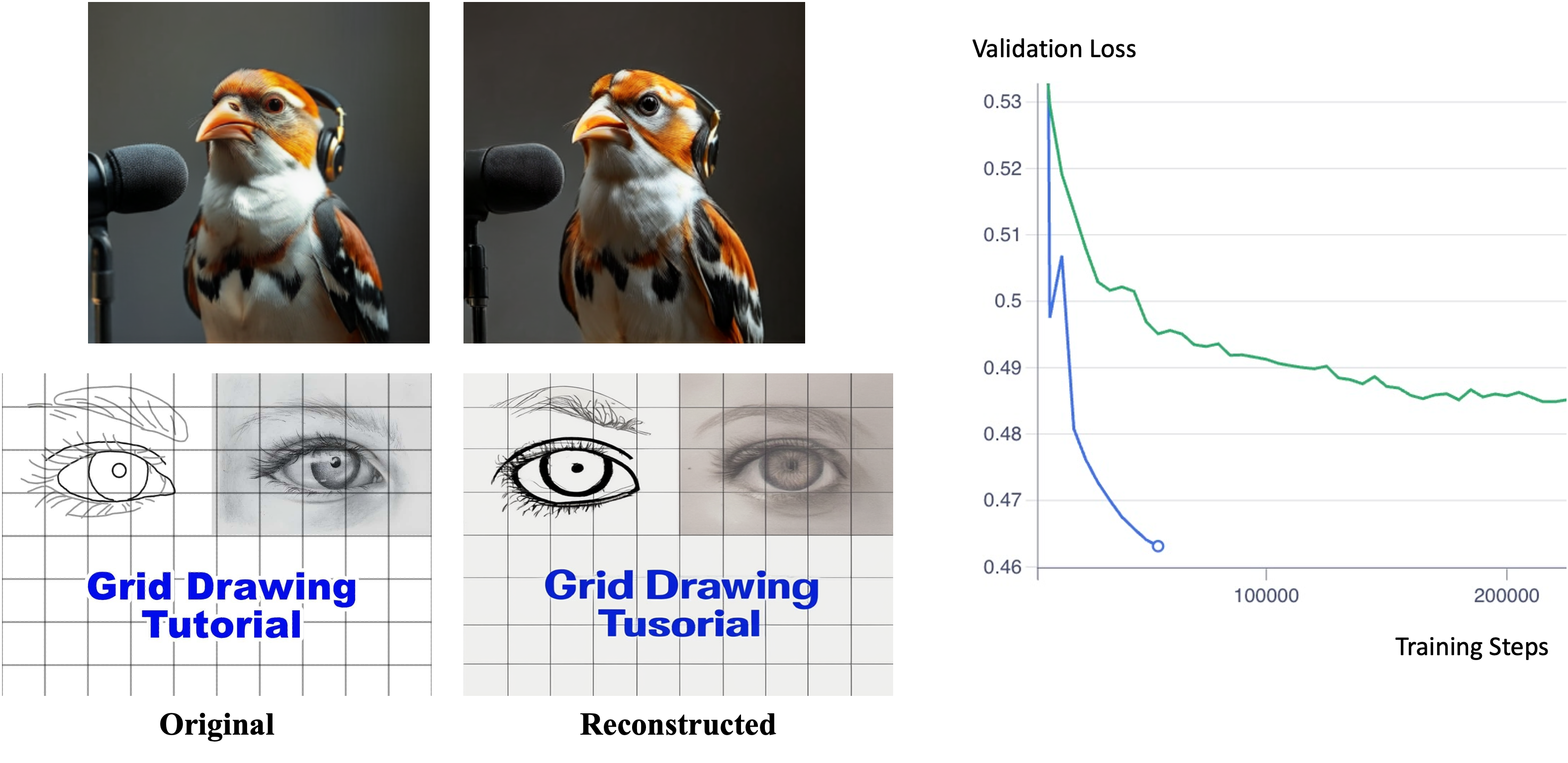}
  \caption{(\textbf{{Left}}) Reconstruction test of TA-Tok~\citep{han_vision_2025} using its accompanying decoder. The reconstruction is imperfect due to unavoidable information loss from semantic abstraction and quantization (see the eye and the feather pattern of the bird, and the tonal difference in both cases). A non-square image shown at the bottom is ``resized to square $\rightarrow$ tokenized $\rightarrow$ decoded as a square image $\rightarrow$ resized back to the original aspect ratio.'' We observe that the distortion level is not critical and presume that it could be compensated for by training a new decoder reflecting this process. (\textbf{{Right}}) Convergence of validation loss for the classic attention-based architecture (green) and our channel-concatenation-based architecture (blue).}
  \label{fig:tatok_recon}
\vskip -0.05in
\end{figure}

\paragraph{Tokenizer.}
We reuse a pretrained text-aligned tokenizer, TA-Tok~\citep{han_vision_2025}, and keep it fully frozen during training. TA-Tok fine-tunes SigLIP~2~\citep{tschannen_siglip_2025} to quantize its output---patch-wise visual features---into discrete tokens and reconstruct the original visual features from these tokens. 
One practical limitation of TA-Tok is its fixed input resolution of 384×384. While the loss of resolution is largely compensated by the diffusion-based vision decoder, non-square images must be resized, which may introduce geometric distortion. We empirically evaluated this issue in advance and found that it does not lead to severe degradation in practice (see Figure \ref{fig:tatok_recon}). The degradation is further mitigated by training our own decoder from scratch, in which such resizing scheme is directly integrated.

\paragraph{Decoder.}

Our vision decoder is similar to the decoders released alongside the TA-Tok model~\citep{han_vision_2025}, but it differs in two key aspects. First, it uses a channel-concatenation-based conditioning architecture that enables significantly faster convergence than attention-based. Second, it supports near-native aspect ratios, avoiding the strict square-image constraint imposed by TA-Tok decoders. 

The model adopts a diffusion transformer composed exclusively of single-stream blocks of MMDiT~\citep{labs_flux1_2025}, 2B parameters in total. It operates on the latent space of FLUX.1 VAE ~\citep{labs_flux1_2025} with patch size 1. Importantly, our model does not use any text-conditioning; the only conditioning signal is vision tokens, which are injected via channel-wise concatenation with the noisy latents. Concretely, discrete vision tokens produced by TA-Tok have a fixed spatial resolution of 27×27; these tokens are first reconstructed into continuous feature vectors and then resized to match the shape of the latents (e.g., 116×78 for a 928×624 image) before concatenation. We empirically observe that this design significantly improves the convergence speed (see Figure~\ref{fig:tatok_recon}). Moreover, by avoiding attention–based conditioning, the overall computational cost of the model is reduced significantly.
Detailed descriptions of the decoder model training and inference are provided in Appendix~\ref{app:vision_details}.

\subsection{Audio Modality}

\hcxoshort is designed to support both audio understanding and generation within a unified language modeling framework. The audio module consists of a continuous audio encoder, a discrete audio tokenizer, and a neural audio decoder. Continuous acoustic embeddings and discrete audio tokens are provided as separate input streams to the language
model, enabling joint processing of audio and text within a single Transformer backbone. For speech synthesis, discrete audio tokens predicted by the language model are passed to the audio decoder, which reconstructs the time-domain waveform.

\paragraph{Encoder.}
For continuous audio representation, we adopt a pretrained audio encoder \citep{Qwen2-Audio}, which is initialized from the Whisper-large-v3 model \citep{radford2023robust}. The input audio is resampled to 16 kHz and transformed into a 128-channel log-mel spectrogram using a 25 ms window size and a 10 ms hop size. A pooling layer with a stride of two is applied to reduce the temporal resolution, such that each output frame approximately corresponds to a 40 ms segment of the original audio. As a result, the encoder produces continuous audio embeddings at an effective frame rate of 25 Hz. Subsequently, the encoder outputs are mapped to the dimension of the language model embeddings via a two-layer MLP adapter consisting of a Linear-GELU-Linear structure. 
To handle audio within video sequences efficiently, we implement an additional token compression mechanism following \citet{kim2025doesaudiomattermodern}. Specifically, we incorporate a single-layer MambaMia module~\citep{kim2025mambamia} after the MLP adapter to further downsample the audio representations from 25 Hz to 1 Hz. This architectural refinement significantly enhances token efficiency, allowing the model to process long-form video-interleaved audio while maintaining a manageable context budget. Throughout the training process, the audio encoder remains frozen to fully leverage the robust acoustic representations learned during large-scale pretraining.

\paragraph{Tokenizer.}

In addition to continuous embeddings, we employ a pretrained audio tokenizer \citep{du2024cosyvoice} to represent speech as discrete units. This tokenizer inserts a finite scalar quantization (FSQ) module \citep{ICLR2024_e2dd5360} into the encoder of a pretrained SenseVoice-Large Automatic Speech Recognition (ASR) model \citep{an2024funaudiollm}. The input speech is first processed by a stack of Transformer blocks to obtain intermediate representations, which are then projected into a low-rank space and quantized using bounded rounding in the FSQ module. The quantized representations are subsequently projected back to the original dimensionality, and discrete audio tokens are obtained by indexing the quantized low-rank vectors in a (2K+1)-ary system. This process yields a codebook of size 6,561 tokens. The resulting audio tokens are generated at a fixed rate of 25 tokens per second, perfectly aligning with the temporal resolution of the continuous audio embeddings.

This dual-encoding design allows the model to exploit the complementary advantages of both representations. Continuous audio embeddings preserve fine-grained acoustic information and rich prosodic details, while discrete audio tokens provide a compact and generation-friendly representation that is well-suited for autoregressive modeling and waveform synthesis.

\paragraph{Decoder.}

To reconstruct time-domain waveforms from discrete audio tokens, we propose an audio decoder named \textit{Unit-BigVGAN}.
The decoder is architecturally derived from BigVGAN-v2 \citep{lee2023bigvgan}, but is adapted to consume discrete audio tokens generated by the LLM rather than continuous mel-spectrogram features. As the decoder directly operates on symbolic unit sequences, which encode limited speaker identity information, the model conditions the generator on an explicit speaker embedding. A reference speech signal is processed by an ECAPA-TDNN \citep{desplanques20_interspeech} to extract a fixed-dimensional speaker embedding that captures speaker-specific characteristics. The embedded discrete tokens are concatenated with the speaker embedding along the channel dimension and used as input to the generator.

Given this combined representation, the decoder follows a BigVGAN-style upsampling and residual processing pipeline to progressively increase temporal resolution and generate the final waveform. The generator consists of an initial convolution layer followed by multiple upsampling stages with residual dilated convolution blocks. Within these residual blocks, a filtered Snake nonlinearity is applied to obtain an anti-aliased representation of discrete-time one-dimensional signals, an approach known as anti-aliased multi-periodicity composition (AMP).
Under this decoding pipeline, each discrete token represents a fixed temporal span of 40 ms, corresponding to a token rate of 25 Hz.

\section{Pre‑Training}

\begin{figure}[t]
\centering
\includegraphics[width=1.0\textwidth]{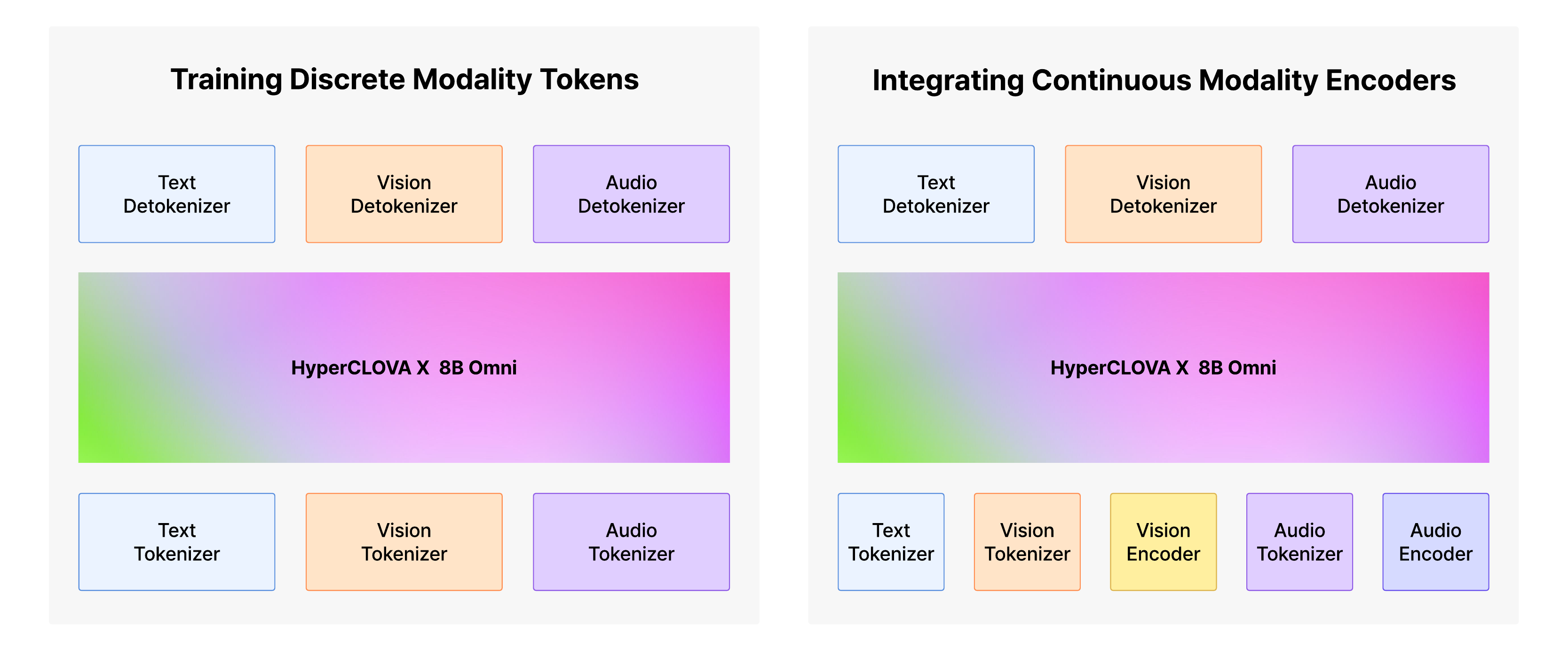}
\caption{Overview of the training process. The model is first trained using discrete modality tokens for text, vision, and audio, establishing a unified symbolic token interface across modalities. Continuous vision and audio encoders are then integrated and jointly trained alongside the discrete tokens, enabling richer multimodal perception within the same Transformer backbone.}
\label{fig:overview_training_process}
\end{figure}

The pre-training consists of two phases to train a text-centric foundation and progressively instill multimodal capabilities.
First, the model learns discrete modality tokens for text, vision, and audio, establishing a unified symbolic token interface that enables joint sequence modeling across modalities. Next, continuous modality encoders for vision and audio are integrated and jointly optimized with the existing discrete representations, allowing the model to incorporate rich perceptual signals while preserving a unified token-based processing paradigm.

As shown in Figure~\ref{fig:overview_training_process}, the training recipe is designed to progressively build up the omnimodal capabilities of the model. Training begins with text-only pre-training to establish a strong sequence-modeling foundation, followed by omnimodal pre-training that extends the backbone to vision and audio through discrete token learning and the integration of continuous-modality encoders. 

\subsection{Text Pre-training}
\paragraph{Data.} The data preparation pipeline follows the scalable preprocessing framework used in \hcxtshort, implemented with a hybrid processing stack based on Datatrove and NeMo-Curator. Raw data are collected and normalized into a unified schema, annotated with document-level quality signals (including PII masking), filtered using a combination of heuristic and model-based criteria, expanded with synthetic data, and finally serialized into sharded files for efficient streaming-based training. Data filtering combines heuristic rules and model-based signals to remove low-quality samples while minimizing unnecessary data loss. This consideration is crucial for Korean corpora, where overall data availability is relatively limited. Filtering policies are pre-defined and applied consistently within each training run. Synthetic data generation leverages two complementary approaches: seed-based generation and document rewriting. Reasoning-oriented synthetic data for STEM, code, and mathematics domains are incorporated during mid-training rather than post-training. The proportion of synthetic data is controlled to preserve training stability while maintaining sufficient diversity. 

\paragraph{Backbone.} The text backbone is pretrained using a multi-stage curriculum with progressively increasing context lengths of $4$K, $8$K, and $32$K tokens. Later stages place greater emphasis on long-form, high-quality, and reasoning-oriented data, with batch sizes adjusted accordingly to accommodate longer contexts. To improve training efficiency under the limited parameter capacity of the $8$B-scale backbone, we employ multi-token prediction~\citep{mtp1}. Specifically, an auxiliary prediction head with a single additional layer is introduced and weighted by a scaling factor of 0.2. This design increases supervision density per token while preserving the original next-token prediction objective, resulting in more effective utilization of training signals without altering the primary optimization target.

\subsection{Training Discrete Modality Tokens}

\paragraph{Stage 1: Multimodal Vocabulary Expansion.}
In this stage, the discrete codebooks produced by modality-specific tokenizers (vision \& audio) are incorporated into the text vocabulary, effectively expanding the model’s token space to support multimodal symbolic representations.
To prevent degradation of text capabilities during the initial multimodal expansion, the text-token portions of the token embedding matrix and the LM head, along with the decoder layers, are frozen, whereas the embeddings for the newly introduced non-text modality tokens are trained for alignment.
This stage is trained for 36K steps, corresponding to 302B tokens, using primarily image--text and audio--text paired data, with text-only data kept at a minimal ratio.
To account for modality token scale differences, the input mixture is controlled with a fixed Image:Audio ratio of 3:1.

\paragraph{Stage 2: Full-Parameter Multimodal Pre-training.}
All parameters are made trainable to enable cross-modal fusion and multimodal reasoning.
This stage performs large-scale end-to-end multimodal pre-training over 2.3T tokens, with modality ratios and loss masking carefully controlled to mitigate text degradation caused by the large vision token budget.
A curriculum-based loss masking strategy is applied to stabilize training. Specifically, during the initial phase spanning the first 1T tokens, the modality mixture is set to Text:Image:Audio = 2:6.5:1.5, with a vision-token loss masking factor of 0.5.
For the second phase, spanning from 1T to 2.3T tokens, the same mixture ratio is maintained while vision loss masking is restored to 1.0.

\paragraph{Stage 3: Long-Context Adaptation for Multimodal.}
A short and focused long-context adaptation is performed to support downstream vision-interleaved and high-difficulty reasoning data under an extended context.
This stage continues from the Stage 2 checkpoint with a 32K context length and a reduced global batch size to improve stability on long sequences, and is trained on approximately 20B tokens.

\subsection{Integrating Continuous Modality Encoders}
In this phase, we integrate continuous modality encoders for both vision and audio to strengthen perceptual modeling and to align continuous and discrete modality representations within a unified sequence modeling framework. While both encoders are incorporated into the architecture, only the vision encoder is actively optimized to enhance visual perception and to align its representations with those produced by the vision tokenizer.

\paragraph{Stage 1: Vision Encoder Alignment.} 
In the first stage, we align visual features with the language model’s embedding space. We keep both the language model backbone and the vision encoder frozen, and we train only a lightweight linear adapter. The training data predominantly consists of image--caption pairs (75.0\%), basic OCR tasks (20.0\%), and VQA samples (5.0\%), establishing a foundational mapping between visual tokens and linguistic representations.

\paragraph{Stage 2: Vision-Centric Full-Parameter Pre-training.}
In the second stage, we train all model parameters to enhance Korean-specific visual perception, including cultural entities, local landmarks, and high-density Korean-script OCR. To preserve previously acquired capabilities across other modalities, we train on a total of 1.5T tokens spanning text (12.1\%), visual understanding (38.5\%), visual generation (34.4\%), and audio (15.0\%). The visual understanding data primarily comprises interleaved in-house and public datasets that capture general visual knowledge, along with text-rich OCR data. The visual generation data includes not only text-to-image samples but also image-editing data. Since image editing utilizes both the vision encoder and the vision tokenizer, this stage further promotes representation alignment between the two components.

\paragraph{Stage 3: Audio Encoder Alignment.}
In the final step of architectural integration, we incorporate the continuous audio encoder to complete the omnimodal framework. While the preceding stages leveraged discrete tokens to accelerate  training, this stage introduces a continuous encoder in parallel to process dense acoustic information. To achieve this, we focus exclusively on ASR tasks, training only a lightweight adapter to bridge the audio encoder and the language model backbone. This stage finalizes the unified architecture, establishing a stable foundation for the subsequent post-training pipeline.

\section{Post-Training}

The post-training is designed to transform the pre-trained omnimodal backbone into a Korean-centric AI assistant capable of seamless, instruction-following interaction across text, audio, and vision modalities. Our primary philosophy centers on a \textit{staged curriculum} that progressively transitions from foundational conversational alignment to complex, intent-aware reasoning. 
While reinforcement learning is recognized as a potent tool for preference alignment, this work primarily focuses on the architectural and data-driven strengths of a supervised fine-tuning (SFT) framework. By establishing this high-fidelity functional baseline, we provide a stable platform capable of seamless cross-modal reasoning, which we identify as the primary milestone for our omnimodal assistant.

\subsection{Data Composition and Strategy}
The post-training data is meticulously curated to balance general linguistic intelligence with specialized multimodal capabilities. We utilize a mix of high-quality human-annotated dialogues and synthetic reasoning traces designed to enhance logical depth. 

A central pillar of our strategy is the prevention of catastrophic forgetting; hence, the curriculum starts with a high concentration of text-based SFT to maintain a stable linguistic foundation before transitioning to complex omnimodal tasks. Specifically, in the initial stage, text-only data constitutes the majority of the training volume ($50.2\%$), ensuring that \hcxoshort preserves its core reasoning and multilingual abilities. 

\begin{figure}[t]
\centering
\includegraphics[width=1.0\textwidth]{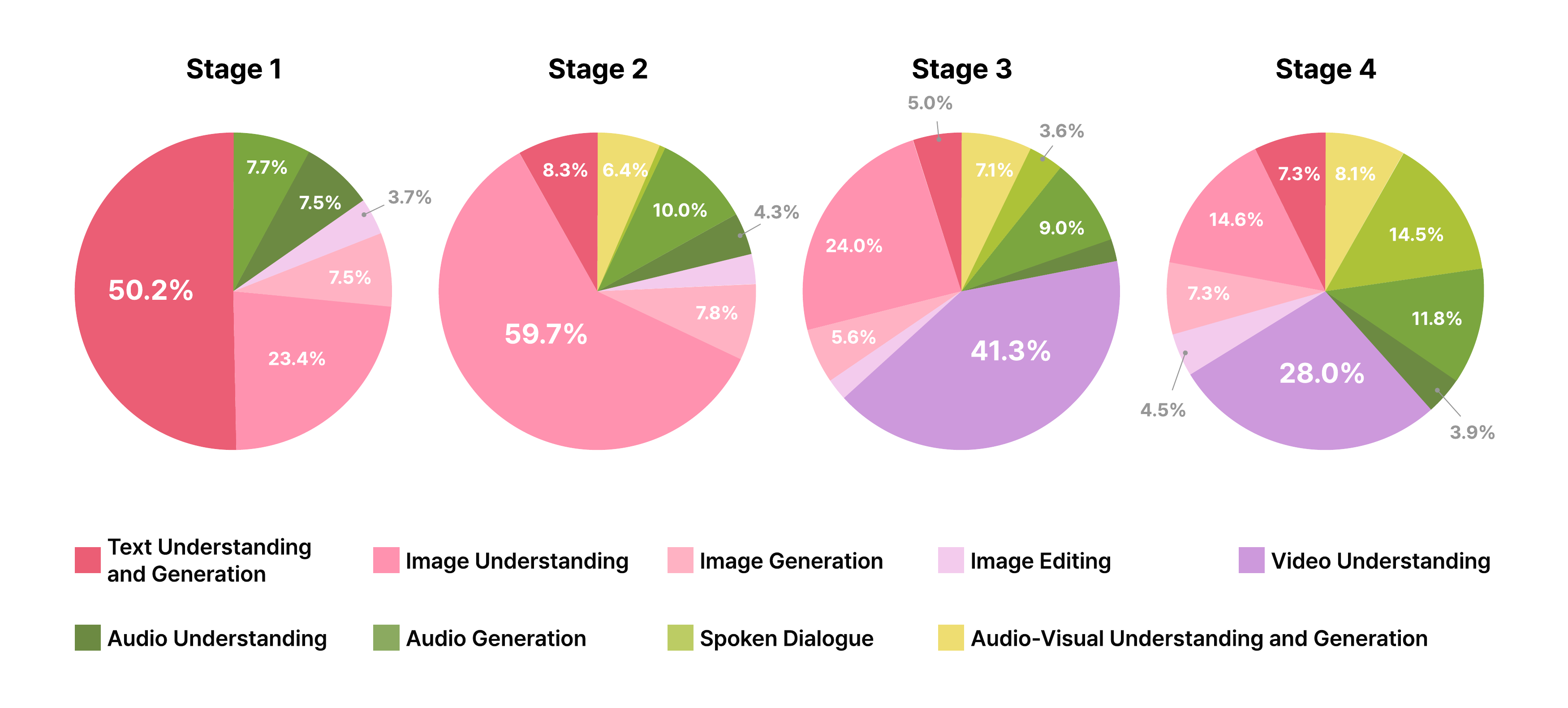}
\caption{Distribution of the post-training datasets across four training stages. Stage 1 focuses on foundational conversational alignment; Stage 2 expands to task-oriented multimodal instructions; Stage 3 introduces temporal and long-context understanding; and Stage 4 refines user-intent reasoning through integrated reasoning paths.}
\label{fig:omni_sft_dist}
\end{figure}

For vision and audio modalities, our strategy emphasizes a progressive increase in task-oriented instructions. This phased, omnimodal-aware approach ensures that \hcxoshort handles diverse any-to-any scenarios without degrading its fundamental capabilities. The detailed distribution of these datasets across the four training stages is visualized in Figure~\ref{fig:omni_sft_dist}.

\subsection{Training Recipe}
The SFT process for \hcxoshort is organized into four sequential stages, each targeting specific functional milestones. 

\paragraph{Stage 1: Foundational Omni Alignment.}
The initial stage is dedicated to establishing foundational instruction-following capabilities by adapting the pre-trained backbone to a conversational framework. As this stage serves as the primary transition point toward a dialogue-centric assistant, it required the most substantial investment of computational resources and training time within the post-training pipeline. Central to this phase is the prioritization of text-based SFT, which constitutes the majority of the training volume ($50.2\%$). This high concentration of linguistic data is critical for ensuring that \hcxoshort maintains a stable linguistic foundation and robust reasoning capabilities before the introduction of more complex omnimodal tasks. This core text-only data is complemented by foundational omnimodal tasks, such as image captioning, ASR, Text-to-Speech (TTS), image generation, and image editing tasks. By allocating the largest portion of the total computational budget to this stage, we ensure a reliable alignment between the model's perceptual outputs and the nuances of human dialogue.

\paragraph{Stage 2: Task-Oriented Omni Specialization.}
The second stage represents a pivotal expansion of the model's functional repertoire, characterized by an exponential increase in both the volume and diversity of instructional tasks. With the linguistic foundation firmly established in Stage 1, the training focus strategically shifts toward large-scale multimodal instruction tuning. During this phase, the proportion of text-only SFT is significantly reduced ($8.3\%$) to accommodate a vast ecosystem of task-oriented omnimodal data, with a primary emphasis on complex image understanding. 
The objective is to cultivate omnimodal synergy by exposing \hcxoshort to heterogeneous scenarios. By navigating these interleaved tasks, the model learns to synthesize cross-modal evidence, enabling it to solve complex queries that require simultaneous processing of text, audio, and vision.

\paragraph{Stage 3: Long-Context and Video SFT.}
Stage 3 is primarily dedicated to temporal modeling and long-context management, with a central focus on video understanding. We integrate a substantial volume of video understanding data ($41.3\%$) and incorporate instruction samples that feature extensive internal reasoning traces to bolster model's logical depth.
This recipe instills \hcxoshort with the ability to maintain semantic coherence over long multimodal sequences and to perform reasoning over temporal events.
To handle the high-resolution audio-visual streams within video efficiently, we introduce a dedicated audio token compressor for video inputs. 
By preceding this stage with a brief alignment phase trained exclusively on video data, with only the compressor module set as trainable, we ensure stable integration of the compressor module into the omnimodal pipeline. 

\paragraph{Stage 4: Intent-Aware Multistep Reasoning.}
The final stage of our post-training curriculum is designed to instill \hcxoshort with the capacity for both high-level intent parsing and sustained, multistep logical reasoning. We internalize a structured reasoning mechanism—the \texttt{<think>} block—to serve as the model's cognitive workspace. 
In this framework, the reasoning trace typically incorporates an initial intent classification step, where \hcxoshort identifies the task category and orchestrates the necessary modality-specific modules. 
For high-complexity tasks, such as STEM problem-solving or cross-modal integration, this initial mapping can naturally transition into an extended deductive reasoning process. 
By systematically breaking down instructions into intermediate logical steps within the latent space, the model can navigate complex problem domains while maintaining strict adherence to user-defined constraints. This versatile reasoning paradigm ensures that \hcxoshort not only selects the correct functional path but also executes deep, context-aware analysis when the task demands it. 
Examples illustrating the foundational intent parsing and task orchestration are provided in Appendix~\ref{appendix:user_intent}.

\section{Evaluation}

\definecolor{col}{HTML}{EEFDE5} 
\newcommand{\NA}{\cellcolor{blue!3}{--}}
\begin{table}[t]
\centering
\setlength{\tabcolsep}{2.3pt}
\scriptsize
\begin{tabular}{ll>{\columncolor{col}}c c c c c c c c}
\toprule
\textbf{Language} & \textbf{Dataset}
& \textbf{\makecell{HyperCLOVA X \\ 8B Omni}}
& \textbf{\makecell{Qwen2.5 \\ Omni 7B}}
& \textbf{\makecell{Emu3 \\ 8B}}
& \textbf{\makecell{Janus-Pro \\ 7B}}
& \textbf{\makecell{X-Omni \\ 7B}}
& \textbf{\makecell{Step-Audio2 \\ Mini 8B}}
& \textbf{\makecell{Qwen2-Audio \\ 7B Instruct }}
& \textbf{\makecell{Audio \\ Flamingo3 7B}} \\
\midrule

\rowcolor{gray!15}
\multicolumn{10}{c}{\textbf{Text-to-Text} $\uparrow$} \\
\midrule

\multirow{4}{*}{Korean}
 & KMMLU-pro        & \textbf{64.9} & 31.1 & 18.7 & 16.4 & 17.7 & 38.6 & 23.8 & 31.8 \\
 & HAERAE & \textbf{75.3} & 51.0 & 20.0 & 14.9 & 21.3 & 59.9 & 40.8 & 46.0 \\
 & KoBALT  & \textbf{27.7}& 17.7 & 10.3 & 5.9  & 9.6  & 17.1 & 8.9 & 12.9 \\
 & Flores+ (En$\rightarrow$Ko) & \textbf{29.2}& 23.7 & 0.5 & 5.7  & 1.3  & 22.7 & 19.2 & 10.5 \\

\midrule

\multirow{4}{*}{English}
 & MMLU        & \textbf{75.7} & 71.6 & 32.4 & 48.2 & 40.0 & 64.9 & 42.9 & 58.7 \\
 & MMLU-pro    & \textbf{54.2} & 50.5 & 10.8 & 20.0 & 19.3 & 40.2 & 17.5 & 30.8 \\
 & GSM8K       & \textbf{87.3} & 87.0 & 3.0  & 43.5 & 61.7 & 75.3 & 39.5 & 62.2 \\
 & Flores+ (Ko$\rightarrow$En) & 27.8 & \textbf{28.6} & 0.9 & 12.1 & 1.2  & 28.2 & 21.6 &18.4 \\

\midrule
\rowcolor{gray!15}
\multicolumn{10}{c}{\textbf{Vision-to-Text} $\uparrow$} \\
\midrule

\multirow{3}{*}{Korean}
 & KoNET      & \textbf{33.0} & 14.7 & 0.6 & 0.3 & 11.3 & \NA & \NA & \NA \\ 
 & K-MMBench  & \textbf{80.2}  & 76.5 & 15.7 & 36.3 & 38.9 & \NA & \NA & \NA \\ 
 & K-DTCBench & 78.8  & \textbf{88.8} & 31.7 & 29.2 & 48.3 & \NA & \NA & \NA \\ 
\midrule

\multirow{4}{*}{English}
 & SEED-IMG & \textbf{80.3} & 77.0 & 69.0 & 72.4 & 74.0 & \NA & \NA & \NA \\ 
 & LLaVA-W  & \textbf{93.8}  & 88.5 & 51.0 & 78.2 & 74.2 & \NA & \NA & \NA \\ 
 & TextVQA  & 80.3  & \textbf{84.4} & 62.9 & 58.7 & 77.5 & \NA & \NA & \NA \\ 
 & DocVQA   & 90.7  & \textbf{94.9} & 74.1 & 43.4 & 88.1 & \NA & \NA & \NA \\ 

\midrule
\rowcolor{gray!15}
\multicolumn{10}{c}{\textbf{Text-to-Vision} $\uparrow$} \\
\midrule

\multirow{2}{*}{English}
 & GenEval & 0.64 & \NA & 0.39 & \textbf{0.78} & 0.67 & \NA & \NA & \NA \\
 & ImgEdit & \textbf{3.83} & \NA & \NA & 1.28 & 1.30 & \NA & \NA & \NA \\

\midrule
\rowcolor{gray!15}
\multicolumn{10}{c}{\textbf{Speech-to-Text (WER $\downarrow$)}} \\
\midrule

\multirow{3}{*}{Korean}
 & KsponSpeech-c
 & \textbf{28.74} & 34.96 & \NA & \NA & \NA & 73.20 & 54.30 & \NA \\
 & KsponSpeech-o
 & \textbf{33.09} & 36.76 & \NA & \NA & \NA & 83.92 & 52.59 & \NA \\
 & Fleurs-ko
 & \textbf{15.33} & 16.23 & \NA & \NA & \NA & 46.20 & 36.86 & \NA \\
 
\midrule

\multirow{3}{*}{English}
 & LibriSpeech-c
 & 1.93 & 4.13 & \NA & \NA & \NA & 11.34 & 3.56 & \textbf{1.41} \\
 & LibriSpeech-o
 & 4.47 & 5.67 & \NA & \NA & \NA & 15.57 & 6.15 & \textbf{3.02} \\
 & Fleurs-en
 & 7.00 & 5.53 & \NA & \NA & \NA & 15.20 & 6.42 & \textbf{3.99} \\

\midrule
\rowcolor{gray!15}
\multicolumn{10}{c}{\textbf{Audio-to-Text (SPIDEr $\uparrow$)}} \\
\midrule

English
 & Clotho-v1
 & 0.259 & 0.051 & \NA & \NA & \NA & 0.238 & 0.138  & \textbf{0.296} \\

\midrule
\rowcolor{gray!15}
\multicolumn{10}{c}{\textbf{Speech-to-Speech (ASR-BLEU $\uparrow$)}} \\
\midrule

En $\rightarrow$ Ko
 & Fleurs-en2ko
 & \textbf{24.70} & 0.00 & \NA & \NA & \NA & 0.09 & \NA & \NA \\

Ko $\rightarrow$ En
 & Fleurs-ko2en
 & \textbf{22.91} & 17.76 & \NA & \NA & \NA & 14.96 &\NA & \NA \\

\bottomrule
\end{tabular}
\caption{Unified Benchmark Results across Text, Vision, and Audio Modalities. For speech-to-text benchmarks, both clean (c) and other (o) splits are used. The symbol `-' indicates that the model does not support the corresponding modality or task. }
\label{tab:unified_all_bench}
\end{table}


To evaluate the performance of \hcxoshort\footnote{In these experiments, we use the checkpoint released under the \texttt{TAG-2025-12-31}, tag: \url{https://huggingface.co/naver-hyperclovax/HyperCLOVAX-SEED-Omni-8B/tree/TAG-2025-12-31}.}, we select a set of open-source multimodal models as baselines, covering text, vision, and audio modalities. The baseline models are chosen based on similarity in model scale, reproducibility of reported evaluation results, and representativeness within each modality. All comparisons are conducted only on the modalities that each model explicitly supports.

Qwen2.5-Omni-7B~\citep{Qwen2.5-Omni} is an Omni model that jointly supports text, vision, and audio, and serves as the most directly comparable baseline to \hcxoshort across all modalities. Both models are based on a general-purpose Omni architecture, enabling consistent comparisons on text understanding and generation, vision–language reasoning, and speech recognition and generation tasks. Accordingly, Qwen2.5 Omni is used as the primary comprehensive baseline throughout the evaluation.

Models such as Emu3 8B~\citep{emu3}, Janus-Pro 7B~\citep{januspro}, and X-Omni 7B~\citep{xomni} are text–vision–centric multimodal models. While they support vision–language understanding and generation, they do not provide audio-related capabilities. Therefore, these models are included as baselines only for text and vision benchmarks.

For audio-related tasks, we include Step-Audio2-Mini-8B~\citep{stepaudio} and Qwen2-Audio-7B Instruct~\citep{Qwen2-Audio} models as baselines. These models can be evaluated on all audio benchmarks. In addition, the Audio Flamingo3 7B~\citep{audioflamingo} is selected due to its reported strengths in English-centric audio and speech-to-text tasks.

Overall, we carefully align each baseline model with the modalities it supports, and apply this principle consistently across both vision and audio evaluations. This evaluation setup allows us to analyze the performance of \hcxoshort as a general-purpose Omni model under realistic and well-defined comparison settings, without relying on unsupported assumptions

\subsection{Text-only Results}

We evaluate the text-to-text performance of \hcxoshort on both Korean and English benchmarks, and the results are reported in Table~\ref{tab:unified_all_bench}.

\paragraph{Korean Text-to-Text benchmarks.}
Across Korean benchmarks, \hcxoshort shows a clear performance advantage over all comparison models. On KMMLU-Pro~\citep{kmmlupro}, HAERAE-1.0~\citep{haerae}, and KoBALT~\citep{kobalt}, \hcxoshort outperforms other large-scale models by a large margin. These results indicate that \hcxoshort effectively learns Korean-based multi-domain knowledge and reasoning abilities, and that the Korean-focused data composition and training strategy directly contributes to performance gains.

\paragraph{English Text-to-Text benchmarks.}
\hcxoshort also achieves strong results on English benchmarks, outperforming comparison models across all evaluation metrics. The MMLU benchmark~\citep{mmlu} series, including MMLU and MMLU-Pro, evaluates broad English multi-task knowledge and reasoning ability. \hcxoshort consistently achieves top-level performance on all benchmarks in this series. In addition, \hcxoshort attains the highest score on the GSM8K~\citep{gsm8k} math reasoning benchmark, showing that numerical reasoning and step-by-step problem-solving abilities are also well learned.

\paragraph{Translation benchmarks.}
We evaluate 1-shot translation performance on the Flores+ benchmark~\citep{nllb-24} using the BLEU metric for the English-Korean pair. For the English-to-Korean direction, we apply Ko-Mecab pre-tokenization to the generated text to compute the scores. As shown in Table~\ref{tab:unified_all_bench}, \hcxoshort achieves the best performance in English-to-Korean translation and delivers performance comparable to the top-performing models in Korean-to-English translation.
These results demonstrate that \hcxoshort has strong cross-lingual capabilities between Korean and English compared to other baseline models.

Overall, \hcxoshort demonstrates strong text-to-text performance on both Korean and English benchmarks compared to existing models. Notably, the performance gap is larger on Korean benchmarks than on English benchmarks, which shows that \hcxoshort provides stable and robust text understanding and reasoning performance in the Korean language setting.

\subsection{Vision \& Text Results}

We evaluate the visual–language understanding and visual generation capabilities of \hcxoshort using a diverse set of public vision benchmarks. The evaluation covers Vision-to-Text tasks, which include image-based question answering and visual reasoning, as well as Text-to-Vision tasks, which focus on text-conditioned image generation and editing. The quantitative results are summarized in Table~\ref{tab:unified_all_bench}.
Furthermore, we extend our evaluation to the temporal domain by assessing the model's video understanding performance on both public benchmarks and specialized internal datasets.
These assessments, detailed in the following paragraph, underscore \hcxoshort's capacity for high-fidelity reasoning across the full visual spectrum.

\paragraph{Vision-to-Text Benchmarks.}
On Korean Vision-to-Text benchmarks, \hcxoshort achieves the best performance on KoNET~\citep{konet} and K-MMBench~\citep{kbench}, and the second-best performance on K-DTCBench~\citep{kbench}. KoNET is a multimodal visual–language reasoning benchmark constructed from a broad range of Korean national educational tests, spanning elementary, middle, high school, and college-level curricula. It evaluates not only the integration of visual and textual information but also diverse forms of Korean linguistic knowledge and educational reasoning across subjects and difficulty levels. The large performance margin observed on KoNET suggests that the Korean-focused data composition and training strategy of \hcxoshort are effective for robust visual–language understanding in the Korean educational context.

\hcxoshort also shows strong performance on English Vision-to-Text benchmarks. It achieves the best results on SEED-IMG~\citep{seedimg} and LLaVA-W~\citep{llavaw}, and the second-best results on TextVQA~\citep{textvqa} and DocVQA~\citep{docvqa}. SEED-IMG and LLaVA-W evaluate general visual question answering and holistic multimodal reasoning, and the strong performance on these benchmarks demonstrates that \hcxoshort effectively aligns visual representations with textual semantics. On benchmarks that emphasize diagram- and document-level understanding, such as AI2D, DocVQA, and ChartQA, \hcxoshort performs slightly below the top model but consistently remains among the top-performing systems, indicating stable and robust visual–language understanding.

\paragraph{Text-to-Vision Benchmarks.}
For Text-to-Vision tasks, we evaluate image generation quality and the accuracy of the text-conditioning capability. \hcxoshort achieves the best performance on ImgEdit~\citep{imgedit}, which focuses on image editing under text constraints, and ranks third on GenEval~\citep{geneval}, which evaluates general text-to-image generation quality. These results suggest that \hcxoshort is particularly strong in preserving semantic intent while performing localized image edits.

An important observation is that \hcxoshort delivers stable performance across both Vision-to-Text and Text-to-Vision tasks while consistently supporting bidirectional text and vision inputs and outputs. In contrast, Qwen2.5 Omni 7B, which performs well on Vision-to-Text benchmarks, does not support Text-to-Vision generation. Janus-Pro and X-Omni support bidirectional text–vision interaction similar to \hcxoshort, but show noticeably lower performance across benchmarks.

\begin{figure}
    \centering
    \includegraphics[width=0.9\linewidth]{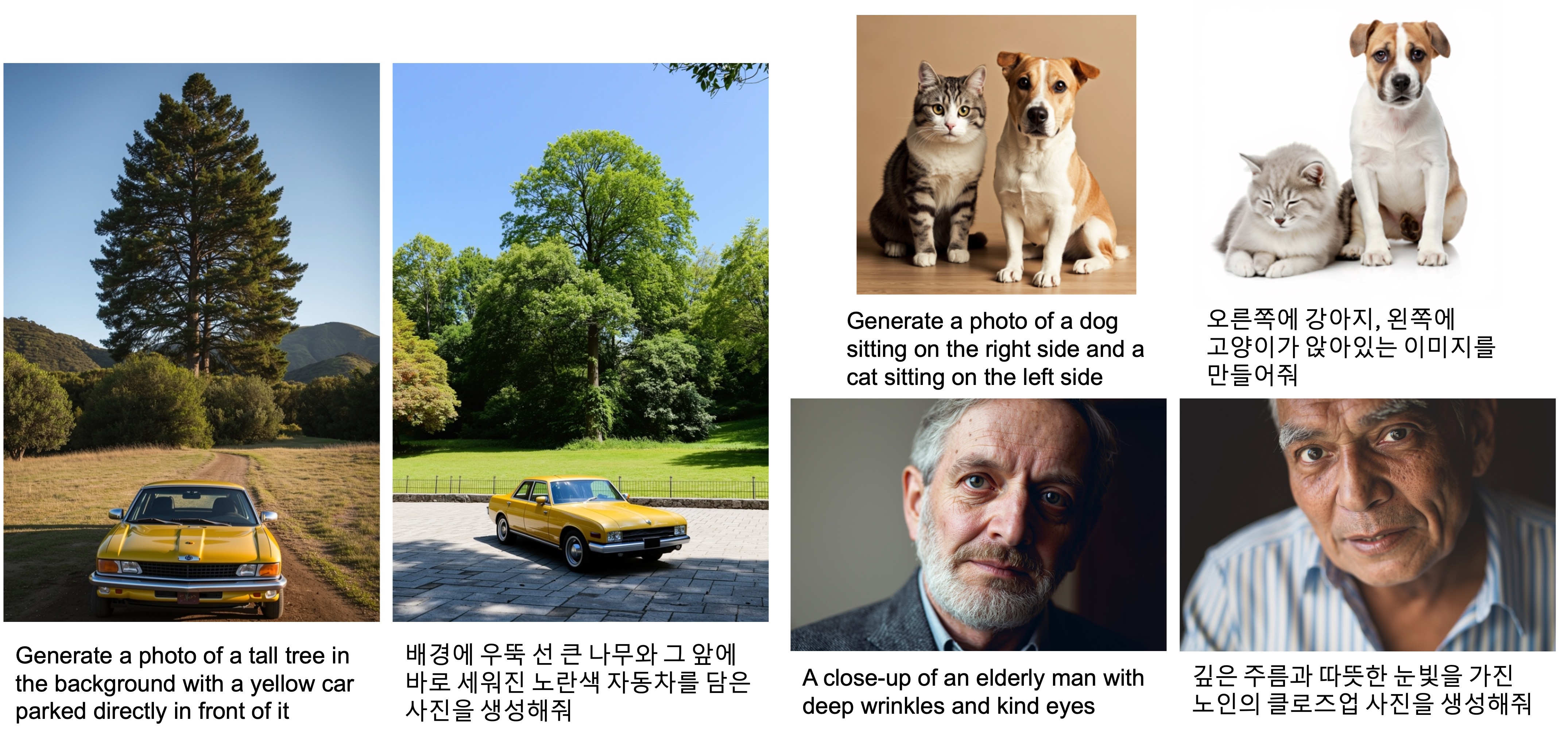}
    \caption{Consistency of generated images from the same semantics, different languages (English and Korean).}
    \label{fig:t2i_en_ko}
\end{figure}

\begin{figure}
    \centering
    \includegraphics[width=0.9\linewidth]{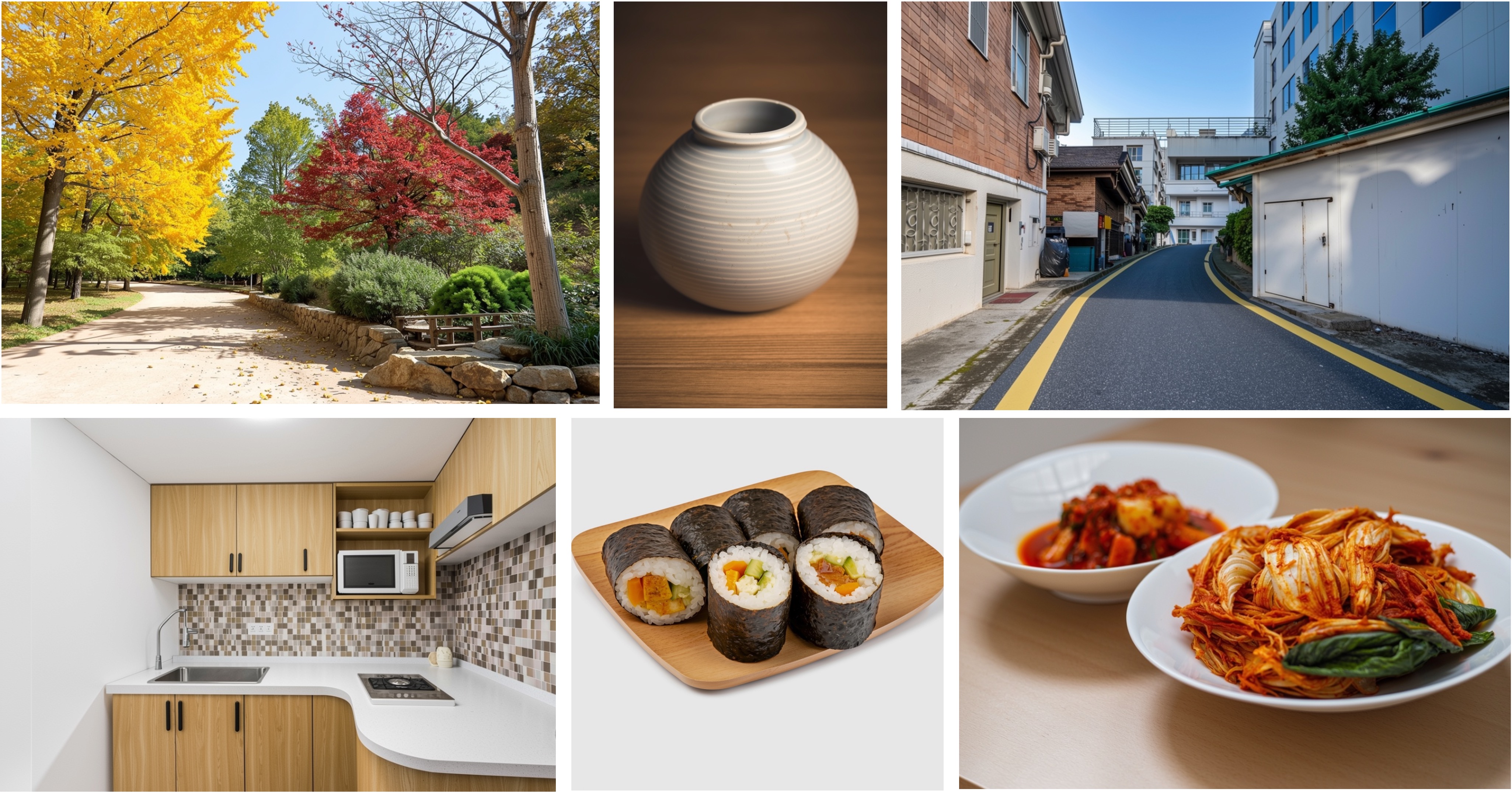}
    \caption{Generated images incorporating Korean cultural attributes.}
    \label{fig:t2i_ko_culture}
\end{figure}

\begin{figure}
    \centering
    \includegraphics[width=0.9\linewidth]{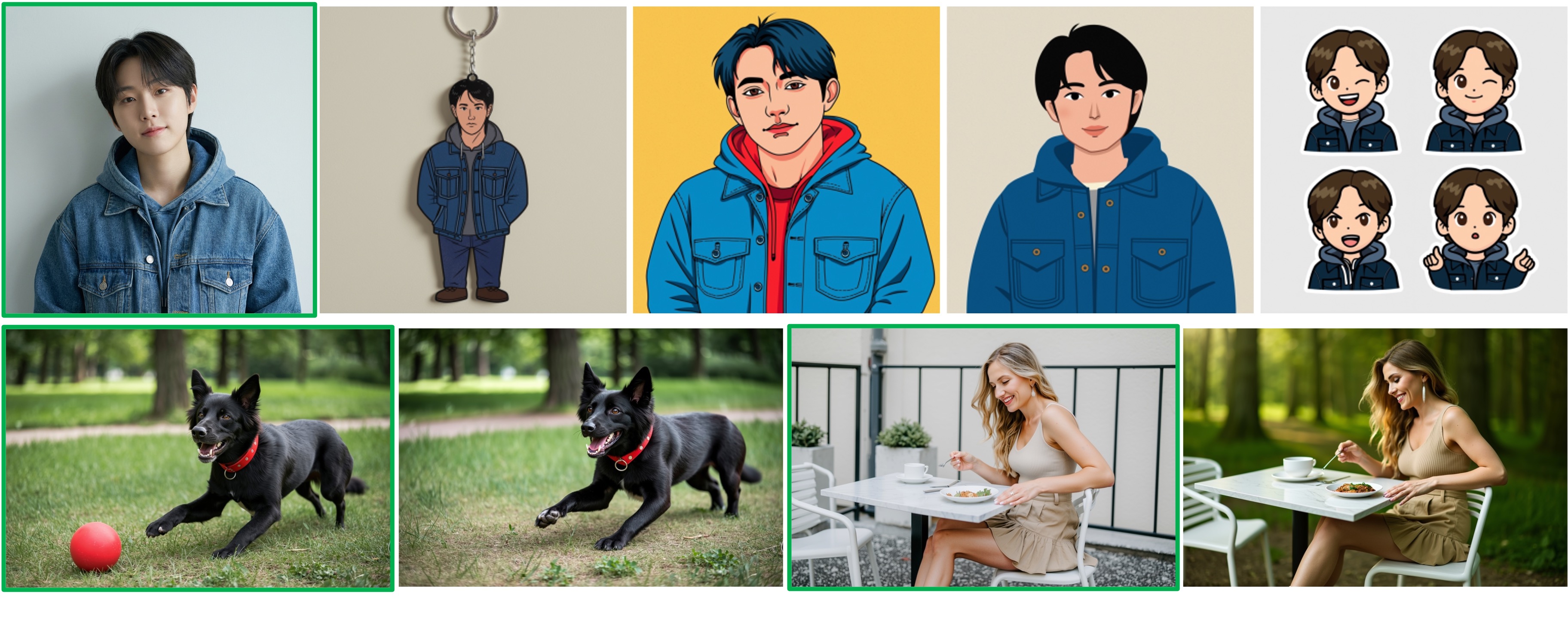}
    \caption{Image editing ability of our model. Input images are marked with green borders. Style change, object removal, background replacement are shown, respectively.}
    \label{fig:edit_demo}
\end{figure}

In addition to quantitative evaluations, we provide qualitative examples to illustrate the text-to-vision capabilities of \hcxoshort. Figure~\ref{fig:t2i_en_ko} demonstrates that the model generates semantically consistent images from prompts expressed in different languages (English and Korean), indicating robust cross-lingual alignment in text-to-image generation. Figure~\ref{fig:t2i_ko_culture} further highlights the model’s ability to incorporate Korean cultural attributes into generated images, reflecting effective grounding in culturally specific visual concepts. Finally, Figure~\ref{fig:edit_demo} showcases the model’s image editing abilities, including style change, object removal, and background replacement, which qualitatively supports the strong performance observed on the ImgEdit benchmark.

Furthermore, \hcxoshort supports not only text and vision but also audio inputs and outputs within a single unified model. This broader multimodal capability provides an additional advantage beyond visual–language tasks and enables more complex multimodal application scenarios. The evaluation of audio-related tasks is presented in the following Section~\ref{sec:audio_bechmarks}.

\paragraph{Video Benchmarks.}
As an omnimodal model built on a unified framework, \hcxoshort natively supports video understanding through its integrated vision--language processing pipeline. We evaluate our model on two distinct benchmarks: Video-MME, a comprehensive multi-modal video evaluation suite, and an internal benchmark assessing NAVER TV~\footnote{\url{https://tv.naver.com/}} comprehension~\citep{navercloud2025hyperclova}, designed to assess the comprehension of real-world Korean video contents. 

The quantitative results are summarized in Table~\ref{tab:video_benchmarks}. In evaluations on Video-MME~\citep{fu2025video} (conducted without subtitles), \hcxoshort achieves a score of 58.2. While this is lower than the 64.3 of Qwen2.5 Omni 7B, it remains highly competitive for an 8B-scale model, comparable to GPT-4V~\citep{openai2023gpt4v} and exceeding LLaVA-NeXT~\citep{zhang2024llavanextvideo} and LLaVA-OneVision\citep{li2025llavaonevision}. Furthermore, on the NAVER TV benchmark~\cite{navercloud2025hyperclova}, \hcxoshort scores 69.7, which is a substantial improvement over GPT-4V’s 50.0. These results indicate that the model's joint training on interleaved Korean-centric multimodal data effectively internalizes the complex temporal and cultural nuances required for specialized video understanding applications.

\begin{table}[t]
\centering
\scriptsize 
\begin{tabular}{ll>{\columncolor{col}}ccccc}
\toprule
\textbf{Language} & \textbf{Benchmark}
& \textbf{\makecell{HyperCLOVA X \\ 8B Omni}}
& \textbf{GPT-4V}
& \textbf{\makecell{Qwen2.5 \\ Omni 7B}}
& \textbf{\makecell{LLaVA-\\OneVision 7B}}
& \textbf{\makecell{LLaVA-\\NeXT 7B}} \\
\midrule

English & Video-MME & 58.2 & 59.9 & \textbf{64.3} & 57.6 & 33.8 \\
Korean & NAVER TV Content & \textbf{69.7} & 50.0 & \NA & \NA & \NA \\

\bottomrule
\end{tabular}
\caption{Performance comparison on video understanding benchmarks. Video-MME results are reported without subtitles. NAVER TV Content is an internal benchmark consisting of real-world Korean video content to evaluate temporal and cultural context understanding.}
\label{tab:video_benchmarks}
\end{table}

\begin{table}[t]
\centering
\scriptsize 
\begin{tabular}{l>{\columncolor{col}}ccccc}
\toprule
\textbf{Language}
& \textbf{\makecell{HyperCLOVA X \\ 8B Omni}}
& \textbf{\makecell{Qwen3-Omni- \\ 30B-A3B}}
& \textbf{\makecell{Gemini{-}2.5 flash}}
& \textbf{\makecell{ElevenLabs v3}}
& \textbf{\makecell{GPT{-}4o{-}mini{-}tts}} \\
\midrule

\rowcolor{gray!15}
\multicolumn{6}{c}{\textbf{Text-to-Speech (MOS $\uparrow$)}} \\
\midrule

English
& 3.94 ($\pm$ 0.14)
& 3.96 ($\pm$ 0.15)
& \textbf{4.44} ($\pm$ 0.14)
& 4.11 ($\pm$ 0.15)
& 4.08 ($\pm$ 0.14) \\

Korean
& \textbf{4.22} ($\pm$ 0.11)
& 3.40 ($\pm$ 0.12) 
& 4.20 ($\pm$ 0.12)
& 4.05 ($\pm$ 0.12)
& 3.43 ($\pm$ 0.12)\\

\bottomrule
\end{tabular}%
\caption{Human Evaluation Results on Text-to-Speech Benchmark. A total of 30 participants evaluated 20 samples per model across 5 models, resulting in 100 evaluations per listener. Values in parentheses represent the 95\% confidence intervals.}
\label{tab:text_to_speech}
\end{table}

\subsection{Audio \& Text Results}
\label{sec:audio_bechmarks}

We evaluate the audio-related performance of \hcxoshort across automatic speech recognition, speech translation, audio captioning, and text-to-speech tasks. Quantitative evaluations based on public benchmarks are reported in Table~\ref{tab:unified_all_bench}, while human evaluations of speech synthesis quality under real-world commercial settings are presented in Table~\ref{tab:text_to_speech}.

\paragraph{Speech/Audio-to-Text Benchmarks.}
On ASR benchmarks, \hcxoshort shows competitive performance in both English and Korean. On Korean ASR benchmarks, including KsponSpeech~\citep{KsponSpeech} and Fleurs-ko~\citep{fleurs}, \hcxoshort achieves state-of-the-art word error rates (WER), establishing strong recognition performance for Korean speech. On English datasets such as LibriSpeech~\citep{librispeech} and Fleurs~\citep{fleurs}, \hcxoshort performs slightly below Audio Flamingo3 7B, but remains competitive with other English speech models.

For audio-to-text tasks, \hcxoshort also achieves the second-highest SPIDEr score on the Clotho-v1~\citep{clotho} audio captioning benchmark, following only Audio Flamingo3 7B, which indicates its ability to effectively summarize acoustic events and semantic information in text form.  We attribute part of these gains to the unified multimodal architecture, which supports generalization across ASR and audio-to-text generation tasks.

\paragraph{Speech-to-Speech Benchmarks.}
We conduct a Speech-to-Speech (S2S) evaluation using a speech-to-speech translation (S2ST) benchmark to evaluate the model's ability to directly convert spoken language from a source tongue into spoken language in a target tongue while maintaining semantic integrity and naturalness. Unlike traditional cascaded systems, \hcxoshort aims to streamline this process, thereby reducing latency and potential error propagation. To rigorously assess these capabilities, we conducted cross-lingual translation tasks between English and Korean using a curated dataset of 270 translation pairs for both En $\rightarrow$ Ko and Ko $\rightarrow$ En directions.

To quantitatively assess the translation accuracy of the generated audio, we utilize the ASR-BLEU metric. This methodology involves transcribing the model's speech output into text, which is then compared against the ground-truth reference via the BLEU score. To ensure a fair and consistent comparison, we employed {\small \texttt{gpt-4o-mini-transcribe}} as the unified ASR engine for all candidate models. BLEU scores were evaluated using sacrebleu~\citep{sacrebleu}. For English-to-Korean translation evaluation, we applied Ko-Mecab pre-tokenization before score computation. Our results demonstrate that \hcxoshort shows superior performance on speech translation tasks; specifically, it achieves the highest performance in both English-to-Korean (En $\rightarrow$ Ko) and Korean-to-English (Ko $\rightarrow$ En) translation directions among all evaluated models.

\paragraph{Text-to-Speech Human Evaluation.} 
To evaluate the performance of Text-to-Speech capabilities, we conducted a Mean Opinion Score (MOS) test focusing on the naturalness of the synthesized speech. A total of 30 human listeners participated in the evaluation. The test set comprised 20 distinct utterances, consisting of 10 English sentences and 10 Korean sentences, to assess the model's proficiency across different linguistic contexts.

The evaluation primarily focused on how closely the synthesized speech resembles human-like pronunciation, intonation, and rhythm. Participants were instructed to rate each audio sample on a 5-point Likert scale, where a score of 1 indicates \textit{"Bad: Due to severe artifacts and lack of naturalness, the audio is nearly unintelligible."} and 5 indicates \textit{"Excellent: The speech is nearly indistinguishable from a real human voice, with natural pronunciation, intonation, and rhythm."}. Details of the evaluation protocol, participant setup, and scoring criteria are provided in Appendix~\ref{appendix:tts_evaluation}.

Table~\ref{tab:text_to_speech} reports the results of human evaluations (MOS) conducted on an internal dataset, comparing \hcxoshort with widely used commercial text-to-speech systems under real-world service conditions. \hcxoshort achieves competitive MOS scores in both English and Korean in terms of naturalness and pronunciation clarity. In particular, \hcxoshort receives higher scores than comparison models for Korean text-to-speech. 

Overall, \hcxoshort achieves balanced performance across a wide range of audio tasks, including speech recognition, audio understanding, speech translation, and text-to-speech. \hcxoshort supports a unified interface in which both inputs and outputs are represented as either audio or text. This design allows the model to be easily extended to multiple audio-related tasks without requiring task-specific model architectures. As a result, heterogeneous tasks such as speech recognition, audio understanding, speech translation, and speech synthesis can be handled in a consistent manner within a single model.

In addition to quantitative evaluations on public benchmarks, human evaluation results conducted under real-world commercial settings show that \hcxoshort achieves quality comparable to existing commercial speech models. In particular, the human evaluation results for text-to-speech indicate that \hcxoshort provides audio quality suitable not only for research settings but also for practical service environments. These results suggest that \hcxoshort, as a general-purpose Omni model, can be effectively applied to a wide range of audio-centric application scenarios.

\section{Conclusion}
In this work, we presented \hcxofull, the first omnimodal model in the \hcx family that supports text, audio and vision modalities as both inputs and outputs. \hcxofull is trained with a unified autoregressive objective that extends next-token prediction beyond text by incorporating discrete vision and audio codebook entries as additional vocabulary items in interleaved sequences. On top of this symbolic interface, continuous vision/audio encoders inject richer perceptual embeddings projected into the same backbone space, while modality-specific decoders translate the shared sequence representations back to pixels and waveforms, compensating for information lost in semantic tokenization.

Empirical evaluations show that \hcxofull achieves competitive performance against comparably sized models across diverse combinations and input and output modalities: text-to-text, vision-to-text, text-to-vision, speech-to-text, audio-to-text, speech-to-speech, and text-to-speech. We expect that its open-sourcing will benefit researchers and practitioners seeking a compact yet versatile model.

The $8$B-scale \hcxofull model serves as the first \textit{pathfinding} point of the design in which a unified auto-regressive backbone supports both interleaved multimodal understanding and any-to-any generation when paired with modality-specific encoders and decoders.
While the performance of \hcxofull is strong relative to its size, 
we anticipate that increasing its size will yield considerable performance gains. A larger and more advanced variant would be particularly valuable in situations with sufficient computational resources where higher performance is desired. Therefore, scaling up the model represents an important avenue for our future research toward developing a collection of robust omnimodal models that can accommodate the needs and restrictions of diverse scenarios.

\bibliographystyle{acl_natbib}
\bibliography{custom,anthology}

\newpage
\appendix

\section*{Contributions and Acknowledgments}
\titlespacing*{\subsection}{0pt}{1ex}{0ex}
\textit{Within each role, \textbf{names are listed in alphabetical order} by first name, followed by the last name.}

\begin{multicols}{2}

\subsection*{Technical Writing}
Cheonbok Park \\ 
Dongyoon Han \\ 
Geewook Kim \\
Hwiyeol Jo \\
Jeonghoon Kim \\
Jin-Hwa Kim \\ 
Jiseob Kim \\
Joonsuk Park\\
Joosung Lee \\ 
Sangdoo Yun \\ 
Sanghyuk Choi \\ 
Sungwook Jeon \\
Taeho Kil \\ 
Yoonsik Kim \\

\subsection*{Model Research and Training}
Bado Lee  \\
Cheonbok Park \\
Daehee Kim \\
Geewook Kim \\
Gichang Lee \\
Hangyeol Yu \\
Heesu Kim \\
Hodong Lee \\
Jeonghoon Kim \\
Jinbae Im \\
Jinhyeon Kim \\
Jiseob Kim \\
Jungwhan Kim \\
Ka Yeon Song \\
Kyeongseok Jeong \\
Moonbin Yim \\
Nako Sung  \\
Ohsung Kwon \\
Sang Hee Park \\
Sanghyuk Choi \\
Seongjin Shin \\
Seunggyu Chang \\
Soyoon Kim \\
Suk Min Seo \\
Taeho Kil \\
Taehwan Yoo \\
Yeontaek Oh \\
Yoonsik Kim \\

\subsection*{Model Evaluation and Analysis}
Daehee Kim \\ 
Dongjin Lee \\
Gayoung Lee \\
Hagyeong Lee \\
Hangyeol Yu \\
Heesu Kim \\
Hwiyeol Jo \\
Hyunhoon Jung \\
Hyunsoo Ha \\
Jeonghyun Lee \\
Jieun Lee \\
Jieun Shin \\
Jonghak Kim \\
Joosung Lee \\
Jungwhan Kim \\
Ka Yeon Song \\
Kiyoon Moon \\
Minkyoung Kim \\
Munhyong Kim \\
MyungIn You \\
Saerim Cho \\
Soyoon Kim \\
Suk Min Seo \\
Taemin Lim \\
Taeyong Kim \\
Woobin Choi \\
Yehbin Lee \\
Yelim Jeong \\
Yeonsun Ahn \\
Yeontaek Oh \\
Youngjin Kwon \\
Zoo Hyun Lee \\

\subsection*{Data}
Byoungeul Kim \\ 
Byungwook Lee \\
Chan-Ho Song \\
Chansong Jo \\
Chiheon Ham \\
Donghyeon Ko \\
Dongjin Lee \\
Eunwoo Song \\
Hanbyul Kim \\
Hoyeon Lee \\
Hyun-Wook Yoon \\
Hyunsoo Ha \\
Injae Lee \\
Jaehong Lee \\
Jaemin Han \\
Jaeuk Byun \\
Jahyeong Lee \\
Jeongmin Liu \\
Jieun Lee \\
Jin-Seob Kim \\
Jinbae Im \\
Jingu Kang  \\
Jisung Wang \\
Jong-Hwan Kim \\
Juncheol Kim \\
Kang Lae Jung \\
Kiyoon Moon \\
Kyeongseok Jeong \\
Min Young Lee \\
Min-Seok Choi \\
Minjae Lee \\
Minkyoung Kim \\
Minseong Choi \\
Moonbin Yim \\
Munhyong Kim \\
MyungIn You \\
Ohsung Kwon \\
Sangkil Lee \\
Seongjin Shin \\
Seunggyu Chang \\
Shinyoung Joo \\
Soo-Whan Chung \\
Sookyo In \\
Soyeon Choe \\
Suhyeon Oh \\
Sung Ae Lee \\
Sungju Kim \\
Sungjun Choi \\
Sunmi Rim \\
Taehong Min \\
Taehwan Yoo \\
Taeyong Kim \\
Yeguk Jin \\
Yehbin Lee \\
You Jin Kim \\
Youna Ji \\
Youngjun Kim \\
Youngki Hong \\

\subsection*{Model Serving and Inference}
Ara Jo \\ 
Eung Sup Cho \\
Gi Nam Kim \\
Hae Song Cho \\
Hanbae Seo \\
Hodong Lee \\
Hyunjoon Jeong \\
Jae Ni Jang \\
Jaeeun Kil \\
Jaegwang Lee \\
Jeongtae Lee \\
Jinhyeon Kim \\
Joonghoon Kim \\
Junhee Yoo \\
Kwang Sun Jung \\
Minjung Jo \\
Minsub Kim \\
Sang Hee Park \\
Sanghyuk Choi \\
Se Hee Lee \\
Seong Chan Hwang \\
Su Jin Lee \\
Sung Bum Hong \\
Sung Woo Lee \\
Sungjae Lee \\
Sungju Kim\\
Ui Seon Yu\\
Woo Sub Kim\\
Yeonsun Ahn \\
Young Beom Lee \\
Young Kyun Jin \\
Young Sang Yoo \\
Young Taek Son \\

\subsection*{Model Planning}
Gayoung Lee \\
Hagyeong Lee \\ 
Hyunhoon Jung \\ 
Jeonghyun Lee  \\
Jieun Shin\\ 
Jonghak Kim \\ 
Saerim Cho\\ 
Taemin Lim \\
Woobin Choi \\ 
Yelim Jeong\\ 
Youngjin Kwon\\ 
Zoo Hyun Lee \\

\subsection*{Business and Brand Strategy}
Dukmin Jung \\
Kyungmin Lee \\ 
Hyojin Park \\
Sujin Roh \\ 
Misuk Park \\

\subsection*{Residency Program}
Bumkyu Park \\ 
Byung Hyun Lee \\ 
Doohyuk Jang \\
Geeho Kim \\
Hyewon Jeon \\ 
Hyunbin Jin \\ 
Hyungwook Choi \\
Ijun Jang \\
Inju Ha \\
Jewon Yeom \\
Jihwan Kim \\
Jihwan Kwak \\ 
Joonki Min \\ 
Juan Yeo \\ 
Junbeom Kim\\ 
Junyeob Kim\\
Kunhee Kim\\
Kyubyung Chae\\
Kyudan Jung\\
Minha Jhang\\
Sangyoon Lee\\
Sehyun Lee\\
Seunghee Kim\\
Song-ha Jo\\
Suho Ryu\\
Yokyung Lee\\

\subsection*{Internships}

Dong-Jae Lee \\ 
Jihwan Moon \\ 
Jinho Heo \\ 
Jisu Jeon \\ 
Minsik Choi \\ 
Seulbi Lee \\ 
Singon Kim \\ 
Sumin Cho \\ 
Woojin Chung \\ 

\end{multicols}


\clearpage
\section*{Appendix}

\section{Implementation Details of the Vision Decoder} \label{app:vision_details}

\paragraph{Training Curriculum.} 
The training of the vision decoder is organized into four sequential phases to stabilize optimization and ensure high-fidelity synthesis across various scales:
\begin{enumerate}
    \item \textbf{Low-resolution crop training}: Initial training at 0.25 Megapixels ($\sim$512$\times$512) focusing on cropped regions.
    \item \textbf{Full-resolution crop training}: Training at 0.6 Megapixels ($\sim$768$\times$768) using 1:2 cropped image regions.
    \item \textbf{Full-resolution full training}: Training on the entire image area at 0.6 Megapixels.
    \item \textbf{Refinement stage}: A final stage with a reduced learning rate to stabilize the model and polish fine visual details.
\end{enumerate}
Crop training is particularly effective in our framework because each cropped region is tightly coupled to its corresponding vision token grid. This allows the diffusion model to unambiguously reference the correct conditioning tokens even when operating on partial image patches.

\begin{figure}[!b]
    \centering
    \includegraphics[width=0.75\linewidth]{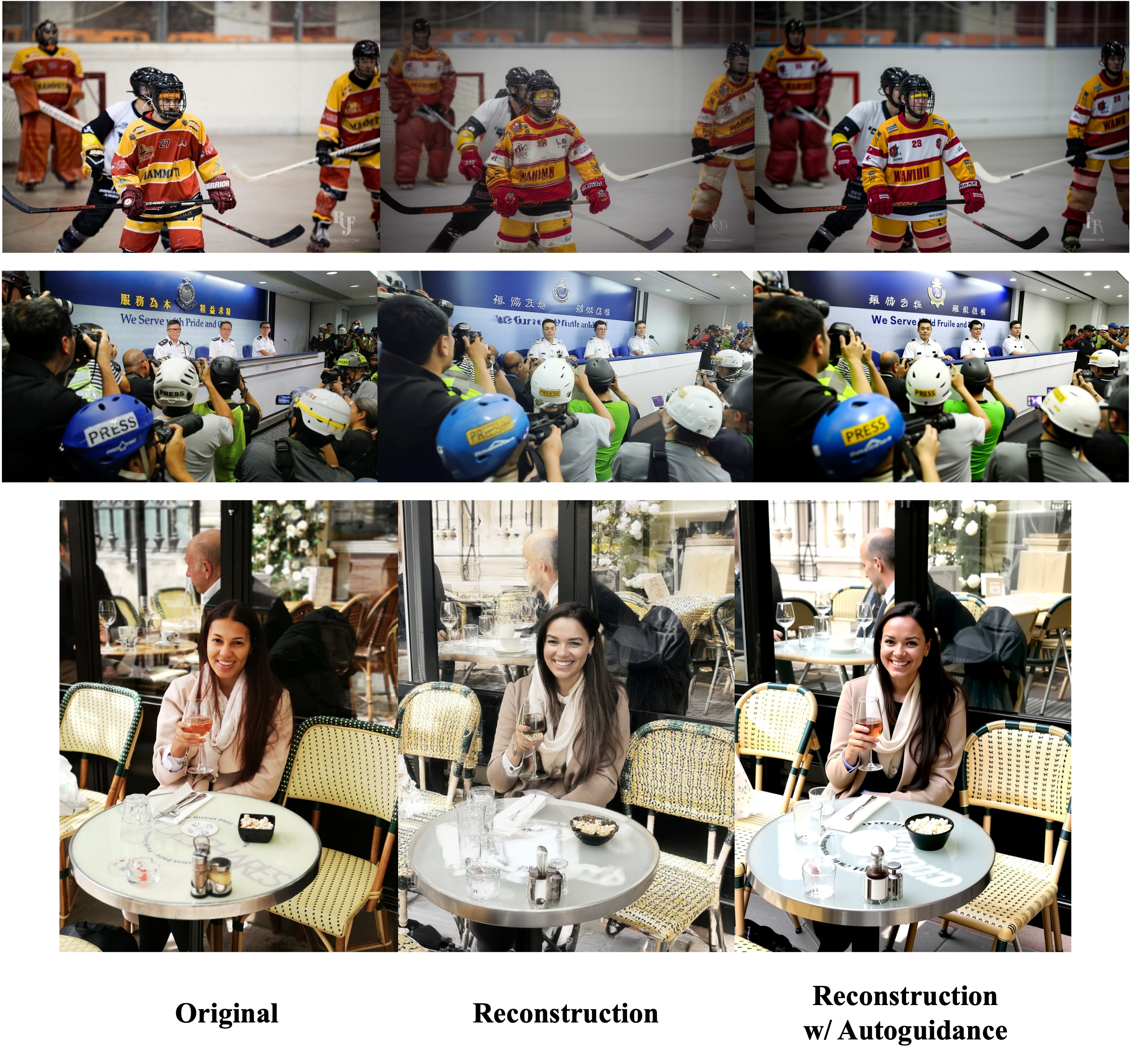}
    \caption{Autoguidance~\citep{karras_guiding_2024} significantly improves the overall quality of our vision decoder (see the enhancement in hockey helmets, typography, and fingers).}
    \label{fig:autoguidance}
\end{figure}

\paragraph{Inference and Autoguidance.}
During inference, we adopt autoguidance~\citep{karras_guiding_2024}, which leads to a substantial improvement in visual quality (see Figure~\ref{fig:autoguidance}). Given that our decoder relies on dense semantic conditioning, the model can occasionally exhibit overly local patterns, resulting in degraded small-scale textures. Autoguidance amplifies the conditioning signal, helping the decoder preserve fine structures—such as typography and intricate patterns—more consistently across the generated output. To apply autoguidance, we train a smaller model (470M) briefly ($\sim$1/20 steps of the main model) and use guidance scale of 1.75.

\newpage
\section{Text-to-Speech Evaluations}
\label{appendix:tts_evaluation}
Participants were presented with synthesized speech from five TTS systems, anonymized and randomly shuffled to prevent model identification. For each text script, annotators listened to all five audio samples and assigned Mean Opinion Scores (1–5) based on fluency and pronunciation clarity. The quality of the synthesized speech was evaluated using a 5-point Likert scale, measuring naturalness and intelligibility. Each score corresponds to the following descriptive criteria:

\begin{description}
    \item[1 (Bad):] Due to severe artifacts and lack of naturalness, the audio is nearly unintelligible.
    \item[2 (Poor):] The audio sounds heavily robotic, and issues with pronunciation or intonation make it uncomfortable to listen to.
    \item[3 (Fair):] While there is a noticeable awkwardness, there are no major issues in understanding the content.
    \item[4 (Good):] The speech is generally natural, though slight awkwardness or a robotic feel may be heard occasionally.
    \item[5 (Excellent):] The speech is nearly indistinguishable from a real human voice, with natural pronunciation, intonation, and rhythm.
\end{description}

\begin{figure}[!b]
  \centering
  \includegraphics[width=1.0\linewidth]{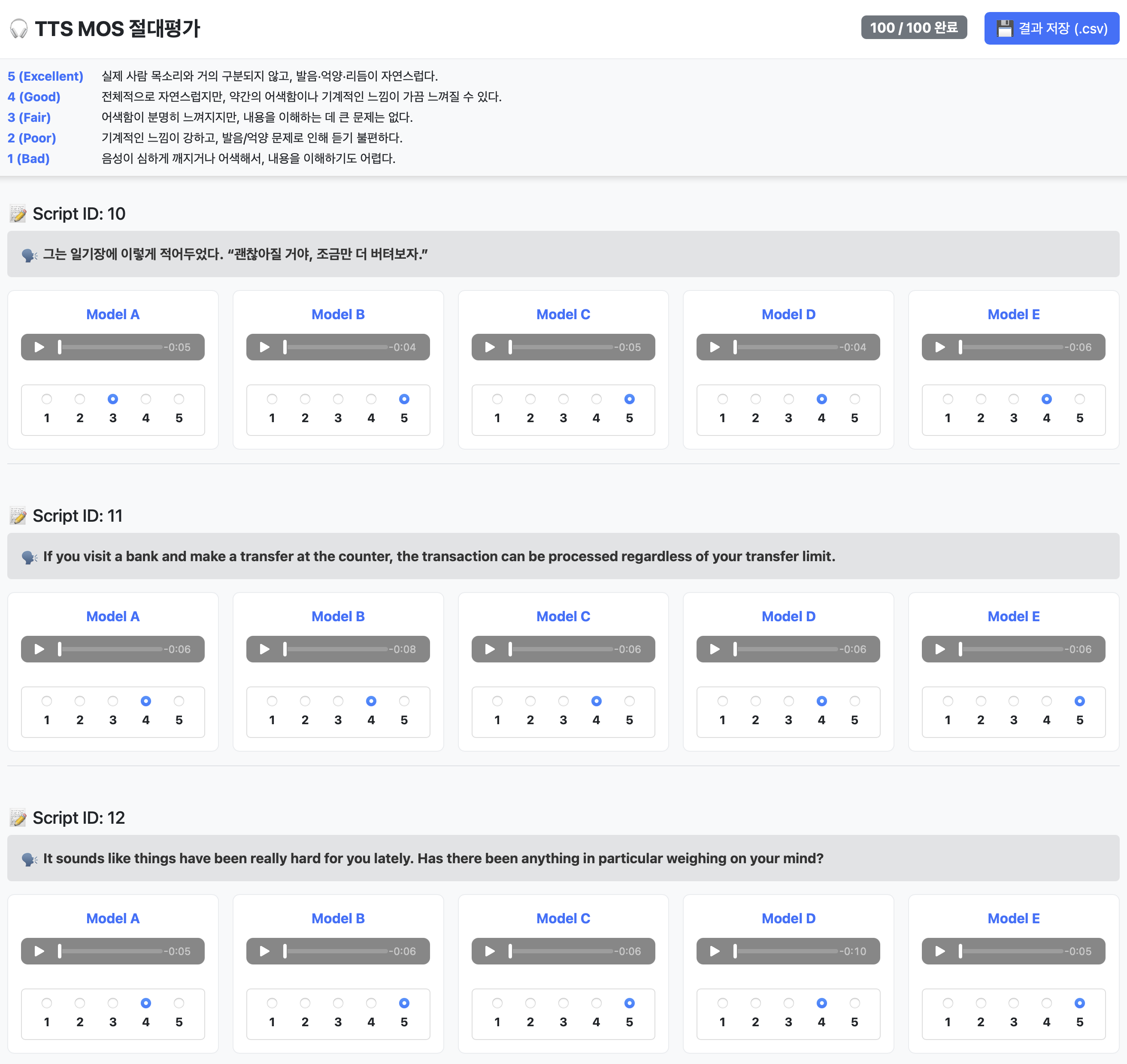}
  \caption{Annotation interface used for MOS evaluation.}
  \label{fig:tts_mos_interface}
\end{figure}

\clearpage
\section{Examples of User-Intent Aware Reasoning}
\label{appendix:user_intent}

As discussed in Section 4.1, \hcxoshort incorporates an internal reasoning process to classify user intent and determine the optimal modality-specific modules for response generation. 
This section provides three illustrative examples of the internalized reasoning (the \texttt{<think>} block), demonstrating the model's ability to parse complex instructions and manage cross-modality constraints.

\subsection{Visual Question Answering}

\begin{quote}
\begin{small}
\begin{verbatim}
    <|im_start|>assistant
    <think>
    This is a VQA task requiring high-level visual reasoning. 
    The input modality consists of an image and a textual query. 
    I will leverage the continuous vision encoder to extract 
    perceptual features and map them to linguistic 
    representations, providing an objective and detailed 
    factual response.
    </think>
    
    [Actual Text Response]
    <|im_end|>
\end{verbatim}
\end{small}
\end{quote}

\subsection{Image Editing}

\begin{quote}
\begin{small}
\begin{verbatim}
    <|im_start|>assistant
    <think>
    This is a request to edit specific visual attributes of 
    the input image. The input modality is image-text
    interleaved. I will utilize the synthesis functions of 
    the diffusion-based vision decoder to modify the requested 
    elements while maintaining the semantic consistency of 
    unchanged objects.
    </think>
    
    [Generated Image Output]
    <|im_end|>
\end{verbatim}
\end{small}
\end{quote}

\subsection{Audio-Visual Understanding with Speech Response}

\begin{quote}
\begin{small}
\begin{verbatim}
    <|im_start|>assistant
    <think>
    This is a complex omnimodal task involving an 
    audio-based user query regarding visual content. 
    I will first process the continuous audio 
    embedding to parse the user's intent, analyze 
    the image features via the vision encoder, 
    and finally generate a synchronized speech response 
    using discrete audio tokens for the neural audio decoder.
    </think>
    
    [Actual Multimodal/Audio Response]
    <|im_end|>
\end{verbatim}
\end{small}
\end{quote}

\end{document}